\documentclass{llncs}
\usepackage[numbers]{natbib}
\usepackage[utf8]{inputenc}
\usepackage{graphicx}
\usepackage{listings}
\usepackage[colorinlistoftodos]{todonotes}
\usepackage[bottom]{footmisc}
\usepackage{xspace}

\let\OldTexttrademark\texttrademark
\renewcommand{\texttrademark}{\OldTexttrademark\xspace}%

\begin{document}

\title{Real-time safety assessment of trajectories for autonomous driving}

\author{
	Hoang Tung Dinh\textsuperscript{1} \and 
	Danilo Romano\textsuperscript{1} \and 
	Patrick Abrahão Menani\textsuperscript{1} \and \\ 
	Victor Vaquero\textsuperscript{1} \and 
	Quentin De Clercq\textsuperscript{1} \and 
	Ramanujan Venkatadri\textsuperscript{2} \and \\ 
	Nelson Quintana\textsuperscript{2} \and 
	Mario Torres\textsuperscript{1}
	}
	
\institute{
	\textsuperscript{1} IVEX NV \\ 
	\email{info@ivex.ai} \\
	\medskip
	\textsuperscript{2} Infineon Technologies AG \\ 
	\email{\{Ramanujan.Venkatadri, Nelson.Quintana\}@infineon.com}
	}

\maketitle

\begin{abstract}
Autonomous vehicles (AVs) must always have a safe motion to guarantee that they are not causing any accidents. In an AV system, the motion of the vehicle is represented as a trajectory. A trajectory planning component is responsible to compute such a trajectory at run-time, taking into account the perception information about the environment, the dynamics of the vehicles, the predicted future states of other road users and a number of safety aspects.

Due to the enormous amount of information to be considered, trajectory planning algorithms are complex, which makes it non-trivial to guarantee the safety of all planned trajectories.
In this way, it is necessary to have an extra component to assess the safety of the planned trajectories at run-time. Such trajectory safety assessment component gives a diverse observation on the safety of AV trajectories and ensures that the AV only follows safe trajectories. 
We use the term trajectory checker to refer to the trajectory safety assessment component. The trajectory checker must evaluate planned trajectories against various safety rules, taking into account a large number of possibilities, including the worst-case behavior of other traffic participants. This must be done while guaranteeing hard real-time performance since the safety assessment is carried out while the vehicle is moving and in constant interaction with the environment. 

In this paper, we present a prototype of the trajectory checker we have developed at IVEX. We show how our approach works smoothly and accomplishes real-time constraints embedded in an Infineon AURIX\texttrademark TC397 automotive platform. Finally, we measure the performance of our trajectory checker prototype against a set of NCAPS-inspired scenarios. 




\end{abstract}


\section{Introduction}

Safety is crucial for autonomous driving. To achieve safety, an autonomous vehicle (AV) must guarantee that its trajectories never cause any accident. Nevertheless, trajectory planning is a complex problem. The trajectory planning component of an AV must take into account an enormous amount of information to compute a trajectory, such as the perception information about the environment, the dynamics of the AV, the predicted future states of other road users, the accomplishment of traffic and safety rules, etc. At the same time, the trajectory planning component must guarantee a low computation time so that it can be used at run-time. 

Due to the complexity of the problem, trajectory planning algorithms are complex with many different steps and techniques integrated, making it non-trivial to guarantee the safety of all planned trajectories. For an overview of the state of the art in trajectory planning algorithms for AV, we refer to \citep{zanchettin2015,paden2016}.

\begin{figure}[t]
	\centering
	\includegraphics[width=\linewidth]{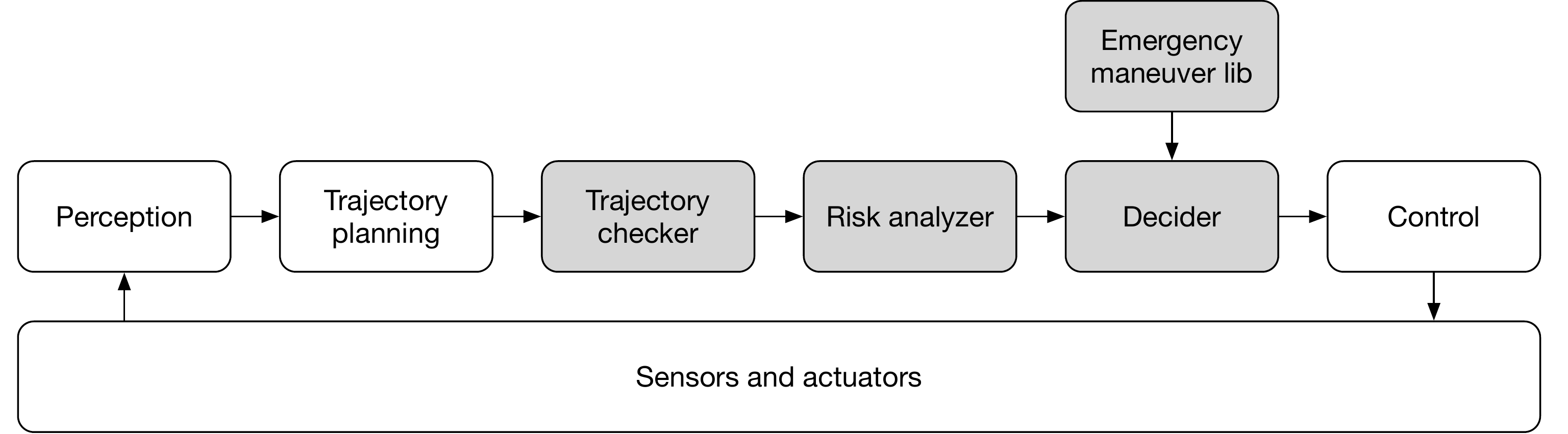}
	\caption{An example architecture of autonomous vehicles with the trajectory checker integrated. The trajectory checker receives candidate trajectories from the trajectory planning components, assesses whether they violate any safety rules and sends the results to the risk analyzer component.}
	\label{fig:Architecture}
\end{figure}

To ensure that an AV will only follow safe trajectories, we propose an additional component to assess the safety of planned trajectories. We call such component a trajectory checker, and it is responsible for evaluating whether the planned trajectories satisfy a set of predefined safety rules or not. Figure \ref{fig:Architecture} shows how a trajectory checker is integrated into the general architecture of an AV.

At each decision-making cycle, the trajectory planning component receives information from the perception component and computes one or several candidate trajectories. The trajectories are then sent to the trajectory checker, which assesses the safety of the candidate trajectories against a set of formalized rules. The safety assessment results are next sent to the risk analyzer component, which computes and weights the risk of each trajectory. Based on the computed risk of each trajectory, the decider component makes a decision on which trajectory the AV should follow. If no candidate trajectory is safe enough to follow, the decider will select an emergency maneuver trajectory from a predefined trajectory library. The selected trajectory is then sent to the control component, which in turn interacts with the actuators of the AV.

The trajectory checker is a crucial component in an AV. It allows different safety rules to be formalized transparently, providing an overview of the safety rules that the AV must respect. At the same time, the trajectory checker provides a hard real-time guarantee on the computation time.

In this paper, we present the first prototype of the trajectory checker that we have developed at IVEX. Since safety rules are often written in natural languages which can be ambiguous, we additionally formalize the safety rules using our proprietary formal language. 
To make sure that the formal specification of the rules is correct before deploying them, its consistency is checked on an iterative process by our solver, the IVEX's solver. If the formalized rules are found to be inconsistent, the solver reports the conflict situations in which there exists no behavior of the AV that could satisfy the safety rules. Finally, once the specification is validated as consistent, the IVEX solver generates a correct-by-construction safety policy in C++, which is the core of our trajectory checker. The trajectory checker can be then deployed and executed on an embedded computer such as the Infineon AURIX\texttrademark TC397 automotive platform to assess the safety of trajectories at real-time.


The trajectory checker prototype has rules of the same nature as the safety rules from the Responsibility-Sensitive Safety paper \citep{shalev-shwartz2017} from Mobileye. It also consists of several additional safety rules defined by IVEX internally. Note that, rules from different resources such as OEMs or safety authorities can be integrated easily into our trajectory checker thanks to the use of IVEX's toolchain during the development.

We benchmark the trajectory checker on a set of scenarios inspired by the European New Car Assessment Programme (EuroNCAP) scenarios. We evaluate the computation performance of the trajectory checker on an Infineon AURIX\texttrademark platform with respect to different trajectory lengths, different numbers of trajectories and different numbers of obstacles. The results show the feasibility of our trajectory checker prototype and its ability to guarantee the hard real-time constraint.

This paper is organized as follows. Section 2 presents a typical architecture of a trajectory checker integrated on an AV. Section 3 describes the development process of our trajectory checker, which is evaluated in Section 4. Finally, Section 5 draws conclusions.

\section{Trajectory checker}

\begin{figure}[b]
	\centering
	\includegraphics[width=\linewidth]{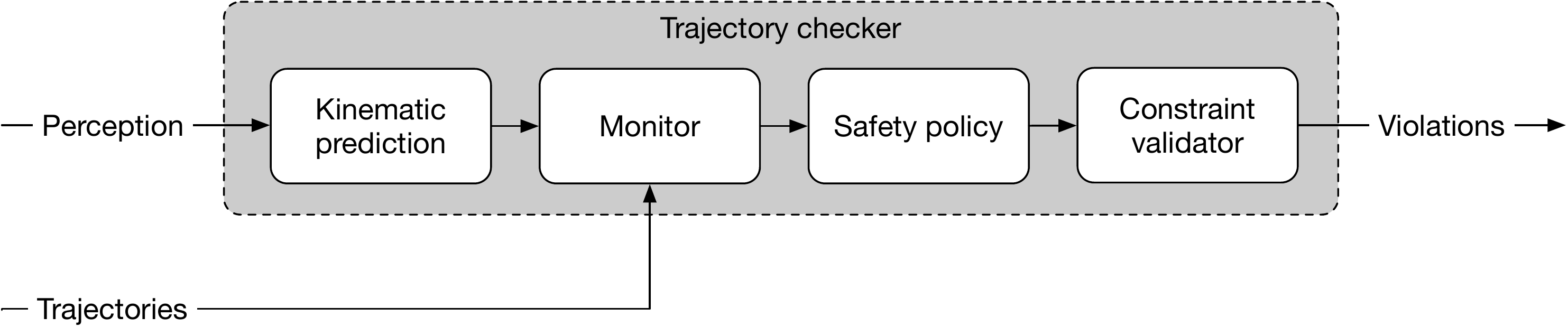}
	\caption{The software architecture of the trajectory checker with 4 modules. The input of the trajectory checker includes perception information and candidate trajectories. The output of the trajectory checker is a list of violations, if there is any, at each trajectory point.}
	\label{fig:TrajectoryChecker}
\end{figure}

We define a trajectory as a list of desired future states of the AV, named trajectory points, through a fixed horizon, for example, 5 seconds. To check the safety of a trajectory, along with the generated trajectory points of candidate trajectories, our trajectory checker receives, possibly redundant, perception information on the state of surrounding objects. As output, we obtain a list of safety violations at each trajectory point. Figure \ref{fig:TrajectoryChecker} illustrates the proposed software architecture of the trajectory checker. 

To perform safety checks on each trajectory point, our trajectory checker firstly predicts the future states of each surrounding obstacle. We perform this using a specific kinematic prediction module. Given a future time associated to a trajectory point, our kinematic prediction module computes, for each obstacle, all the possible states based on a set of assumptions about their worst-case behaviors.


Following Figure~\ref{fig:TrajectoryChecker}, in the next step both the predicted states of obstacles and the checked trajectory are discretized to semantically meaningful information by the monitor module. The discrete information is then used by the safety policy module which contains all the encoded safety rules. Based on its inputs and safety rules, this module decides which motion constraints must be respected by each trajectory point. Finally, given the motion constraints provided by the safety policy, the constraint validator module checks which ones are violated by the trajectory points, outputing them as violations.

\section{Safety policy}

At the heart of the trajectory checker is a safety policy. At run-time, the safety policy decides which motion constraints a trajectory must satisfy given the current situation. 
Developing a safety policy is non-trivial. On one hand, it needs to take into account many possible situations that might occur. 
On the other hand, it needs to correctly enable motion constraints per situation concerning the safety rules.

\begin{figure}[b]
	\centering
	\includegraphics[width=0.7\linewidth]{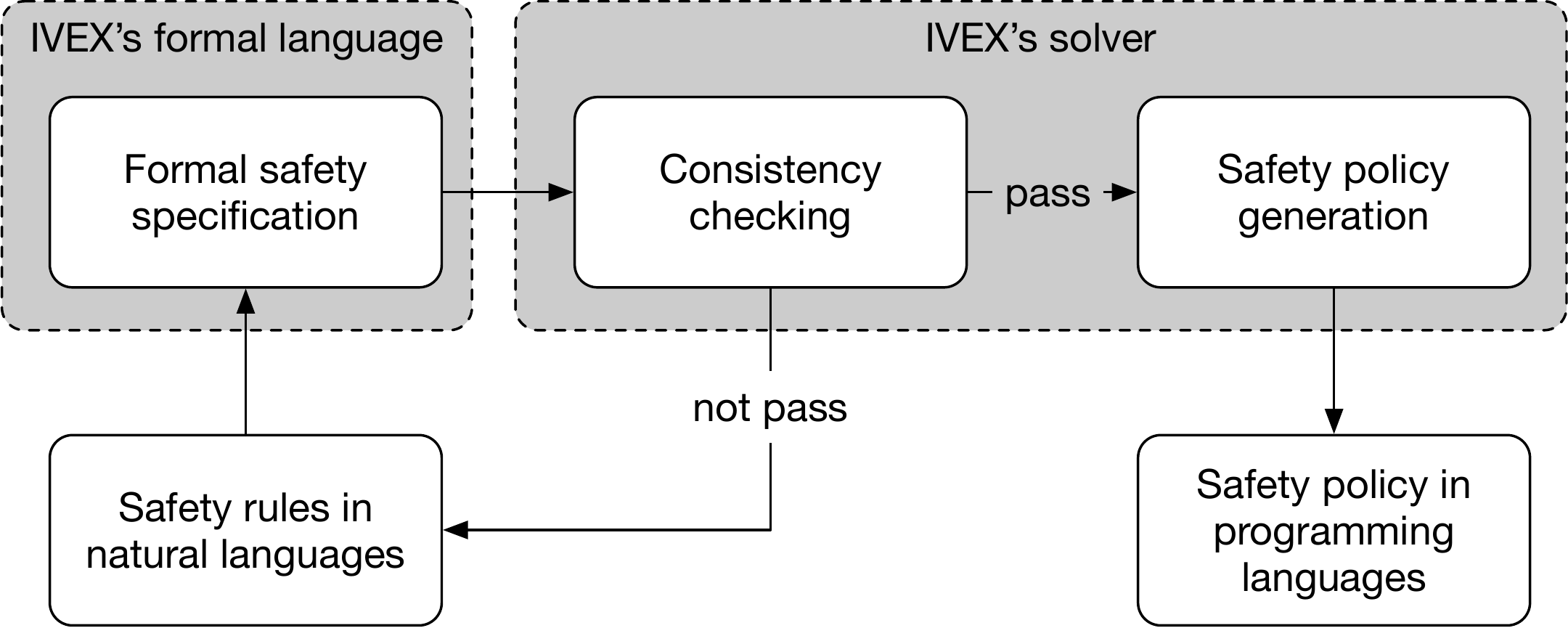}
	\caption{Safety policy development process. The process is iterative and makes use of IVEX's formal language and IVEX's solver. The input of the process is a set of safety rules in natural languages. The output of the process is a generated safety policy in programming languages, for example, C++.}
	\label{fig:Process}
\end{figure}


In this section, we describe the development process of a safety policy starting from a set of safety rules written in natural languages, as illustrated in Figure \ref{fig:Process}. To guarantee the correct implementation of the safety policy, we use different formal approaches in this development process.

First, given a set of safety rules defined in natural languages, we formalize them using a formal language developed at IVEX. The formal safety specification is then checked for consistency by IVEX's solver. If the safety specification is found to be inconsistent, the situations in which any inconsistency happens are reported to the developers, so that they can inspect and correct the safety rules. Once the safety specification is consistent, the IVEX's solver generates a correct-by-construction safety policy in C++.

IVEX's formal language is based on first-order logic, which is declarative. This is a fundamental aspect of the approach and makes the specification of safety rules more manageable and adaptable. 
Our formal language represents the capability of the AV systems by using a specific construct named \emph{action}. The environment and the internal states of the AV are represented by a set of discrete state variables. 
The language allows the formalization of safety rules as goals based on the value of the discrete state variables and the actions. It also allows the specification of the preconditions and the expected effects of each action, as well as mutual exclusive constraints among actions. Safety rules can be prioritized in the language as well as can be enforced in parallel. 
More details on the formal language and a concrete safety model will be published in a follow-up paper.


IVEX's solver uses advanced techniques in formal verification and artificial intelligence such as automated planning, constraint satisfaction and model checking to inspect the consistency of a safety specification. Consistency means that there exists no situation in which two safety rules are in conflict so that no action can be taken to avoid a safety violation. For example, if one safety rule requires the vehicle to change to the left lane while another safety rule requires the vehicle to change to the right lane, it is impossible for the autonomous vehicle to satisfy both safety rules at the same time. If there exists a situation, the safety specification is inconsistent. It means that there is a mistake in the specified safety rules to be revised by the safety engineers.

Once the safety specification is validated as consistent, the IVEX's solver generates a safety policy in the C++ programming language. The generated safety policy is guaranteed to be correct-by-construction with respect to the formalized safety rules. IVEX's solver also uses advanced techniques such as logic minimization to optimize the computational performance of the safety policy.

The safety policy is a tree-like data structure with a know maximum depth, which is critical for guaranteeing the hard real-time execution requirement. As input to any safety policy, we have the values of the discrete state variables provided by the monitor module. The safety policy then maps each possible situation, defined in terms of its inputs, to a set of safety constraints to be respected by the trajectory.

\section{Kinematic prediction}

To assess the safety of a trajectory, the future states of all the surrounding objects must be predicted. We use a conservative approach to compute the possible future states of these objects based on assumptions on their worst-case behavior. 
As a non-limiting example, the current prototype includes the following assumptions:

\begin{itemize}
	\item The maximum and minimum longitudinal velocity
	\item The maximum and minimum lateral velocity
	\item The maximum acceleration
	\item The maximum braking deacceleration
\end{itemize}

Given the assumptions and the current state of each object, the kinematic prediction module computes its future worst-case position and velocity. The information is then used by the safety policy to decide which safety constraints must be respected by the trajectory.

\begin{figure}[b]
	\centering
	\includegraphics[width=0.6\linewidth]{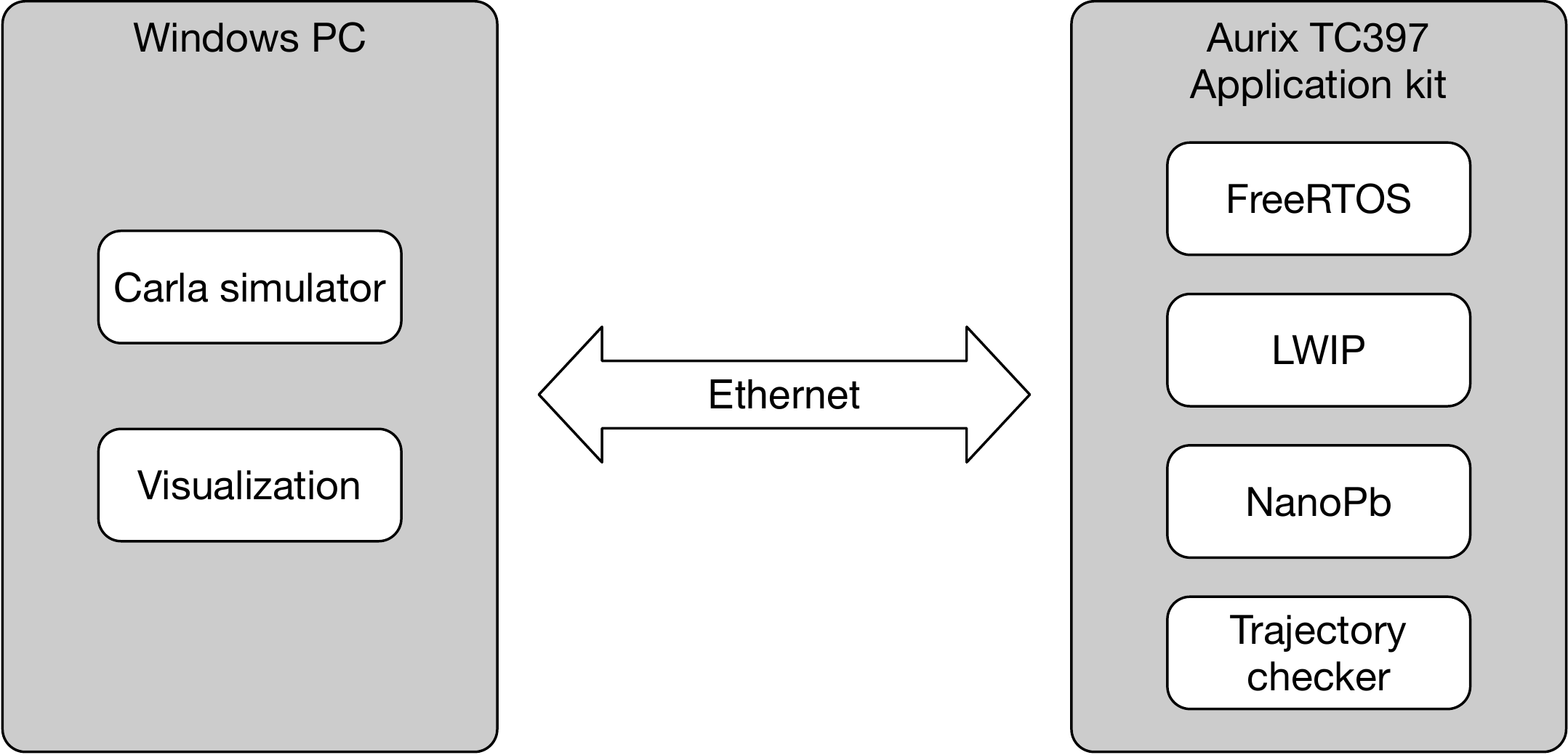}
	\caption{Experiment setup. The trajectory checker is executed on the AURIX\texttrademark TC397 Application kit together with other libraries for communication. The Carla simulator runs on a Windows PC and communicates with the AURIX\texttrademark kit via an Ethernet connection.}
	\label{fig:ExperimentSetup}
\end{figure}

\section{Validation and benchmarking}

We execute the developed trajectory checker on the Infineon AURIX\texttrademark TC397\footnote{https://www.infineon.com/cms/en/product/evaluation-boards/kit\_a2g\_tc397\_5v\_tft/} automotive platform to validate, measure and benchmark its performance. 
The AURIX\texttrademark TC397 is a high-performance and safety microcontroller specifically suitable for automotive applications. 
We benchmark the execution of the trajectory checker against 24 scenarios created based on the EuroNCAP scenarios. These benchmarking scenarios include a variety of safety-critical traffic situations such as cut-in, cut-out, sudden stop and traffic jam.

\subsection{Experiment setup.}

Figure \ref{fig:ExperimentSetup} describes our experiment setup. The testing scenarios are run by the Carla \citep{Dosovitskiy17} simulator on a Windows PC. The trajectory checker is running on the AURIX\texttrademark TC397 application kit. The Windows PC and the AURIX\texttrademark application kit communicates via an ethernet connection. We use Google's Protobuf as the message exchange format between the Windows PC and the AURIX\texttrademark application kit.

There are four software components running on the AURIX\texttrademark application kit, including FreeRTOS - an operating system, LWIP - a driver for lightweight IP communication, NanoPb - a library to serialize and deserialize Protobuf messages and the trajectory checker. All software components run on a single core (cpu0) of the AURIX\texttrademark application kit and are compiled by the Hightec compiler version v4.9.3.0.

We use as ground-truth the perception and localization data provided by Carla. Every 100 milliseconds, this ground-truth information from Carla, as well as the future trajectory of the AV, are sent from the Windows PC to the AURIX\texttrademark application kit as a Protobuf message. After the application kit receives the data, the received Protobuf message is decoded and the trajectory checker is invoked to perform the safety assessment. Once the results are ready, they are sent back to the Windows PC for visualization and data collection.

\subsection{Scenario description}

The scenarios in our testing set are inspired by the NCAPS scenarios. We specify them in OpenSCENARIO format and are executed in the Carla simulator. All the scenarios take place in a straight road with four lanes, numbered from 0 to 3 from right to left
This default setting was chosen to resemble highway driving conditions. We now explain the details of some scenarios.

\begin{itemize}
	\item \textbf{Cut-in}
	
	This scenario is designed to validate the predictive capability of the trajectory checker.
	In this scenario, two vehicles are involved: the ego vehicle and a guest vehicle. 
	At the beginning of the scenario, the ego vehicle is driving along lane 2, while the guest vehicle is driving on lane 3.
	During the scenario execution, the guest vehicle suddenly moves to the lane of the ego vehicle, generating the cut-in situation.
	Three different scenario variants are considered based on the distance between the ego vehicle and the guest vehicle at the cut-in moment.

	\item \textbf{Cut-out}

	This scenario is designed to validate longitudinal velocity and acceleration safety rules of the trajectory checker.
	In this scenario three vehicles are involved: the ego vehicle, a moving guest vehicle and a stopped guest vehicle.
	At the beginning of the scenario, all the vehicles are on lane 2. The moving guest vehicle is in between the stopped vehicle and the ego vehicle.
	During the scenario execution, the ego vehicle and the moving guest vehicle drive towards the stopped vehicle. When the guest moving vehicle approaches the stopped vehicle, it suddenly changes its lane to avoid the stopped vehicle.
	Three different scenarios are considered based on the distance between the moving guest vehicle and the stopped vehicle at the moment the moving vehicle changes lane.

	\item \textbf{Traffic jam}
	
	This scenario is designed as a stress test for measuring how the trajectory checker scales with the number of objects.
	In this scenario, 31 vehicles are involved: the ego vehicle and 30 guest vehicles.
	At the beginning of the scenario, the ego vehicle is driving along lane 2 and all the vehicles are distributed over the remaining lanes.
	During scenario execution, all the vehicles move straight-forward. At a specific moment, one guest vehicle changes from lane 1 towards lane 2, ending up just in front of the ego vehicle.
	This scenario is similar to the cut-in scenarios but the ego vehicle will need to consider a larger number of guest vehicles in the scene.
	
	\item \textbf{Lane change due to obstacles}
	
	This scenario is designed to validate lateral velocity and acceleration safety rules of the trajectory checker.
	In this scenario two vehicles are involved: the ego vehicle and a stopped guest vehicle.
	At the beginning of the scenario, the two vehicles are both on lane 2. The stopped vehicle is in front of the ego vehicle.
	During the scenario execution, the ego vehicle starts approaching the stopped vehicle. The ego vehicle needs to perform a lane change to avoid the stopped vehicle. 
	
	

	\item \textbf{Double cut-in}
    
	This scenario is designed to validate the trajectory checker when two dangerous maneuvers happen at the same time.
    In this scenario, three vehicles are involved: the ego vehicle and two moving guest vehicles.
    At the beginning of the scenario, the ego vehicle is driving along lane 2, while the other two vehicles are on lanes 1 and 3, respectively. 
    During the scenario execution, the ego vehicle drives straightforward. At a specific moment, both the two guest vehicles change to the ego vehicle's lane at the same time.

\end{itemize}

\begin{figure}[t]
	\centering
	\includegraphics[width=0.5\linewidth]{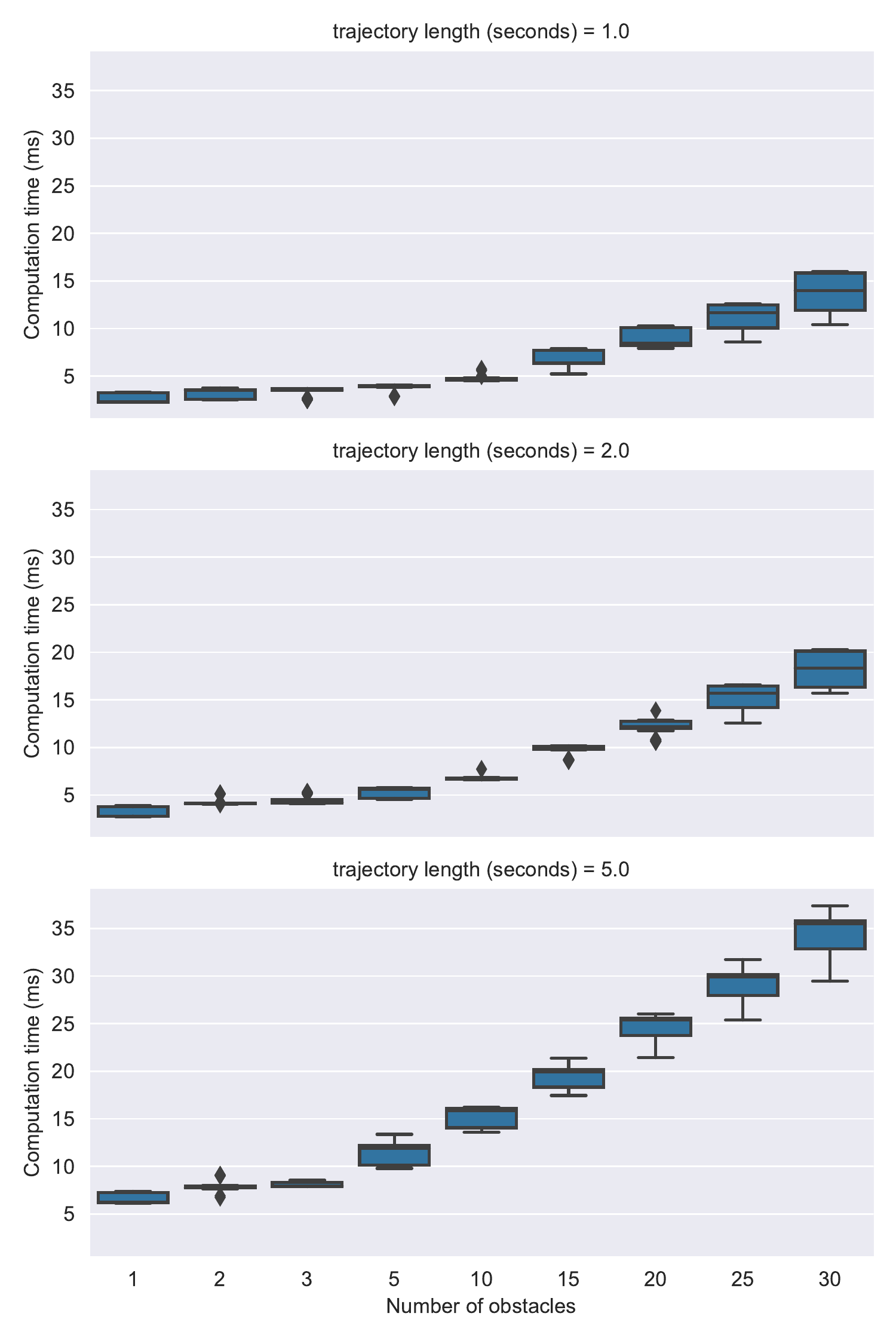}
	\caption{The computation time of the trajectory checker on IVEX's scenarios.}
	\label{fig:ComputationTime}
\end{figure}

\subsection{Benchmarking results}

We measure the computation time of the trajectory checker against different input sizes, defined in terms of the length of the trajectory and the number of the objects. We only measure the time that the trajectory checker is invoked, not accounting for the communication and the message serialization/deserialization times.

Figure \ref{fig:ComputationTime} presents the computation time of the trajectory checker for a single trajectory. The results show that the computation time of the trajectory checker increases linearly with respect to the number of obstacles. When dealing with one-second trajectories, the trajectory checker can perform a safety assessment in situations with 30 obstacles in around 15 milliseconds. For two-second trajectories, the computation time when there are 30 obstacles increases to about 20 milliseconds. Finally, it can be observed that for a five-second trajectory and 30 obstacles, it takes less than 40 milliseconds for the trajectory checker to perform the safety assessment.

\begin{figure}[t]
	\centering
	\includegraphics[width=\linewidth]{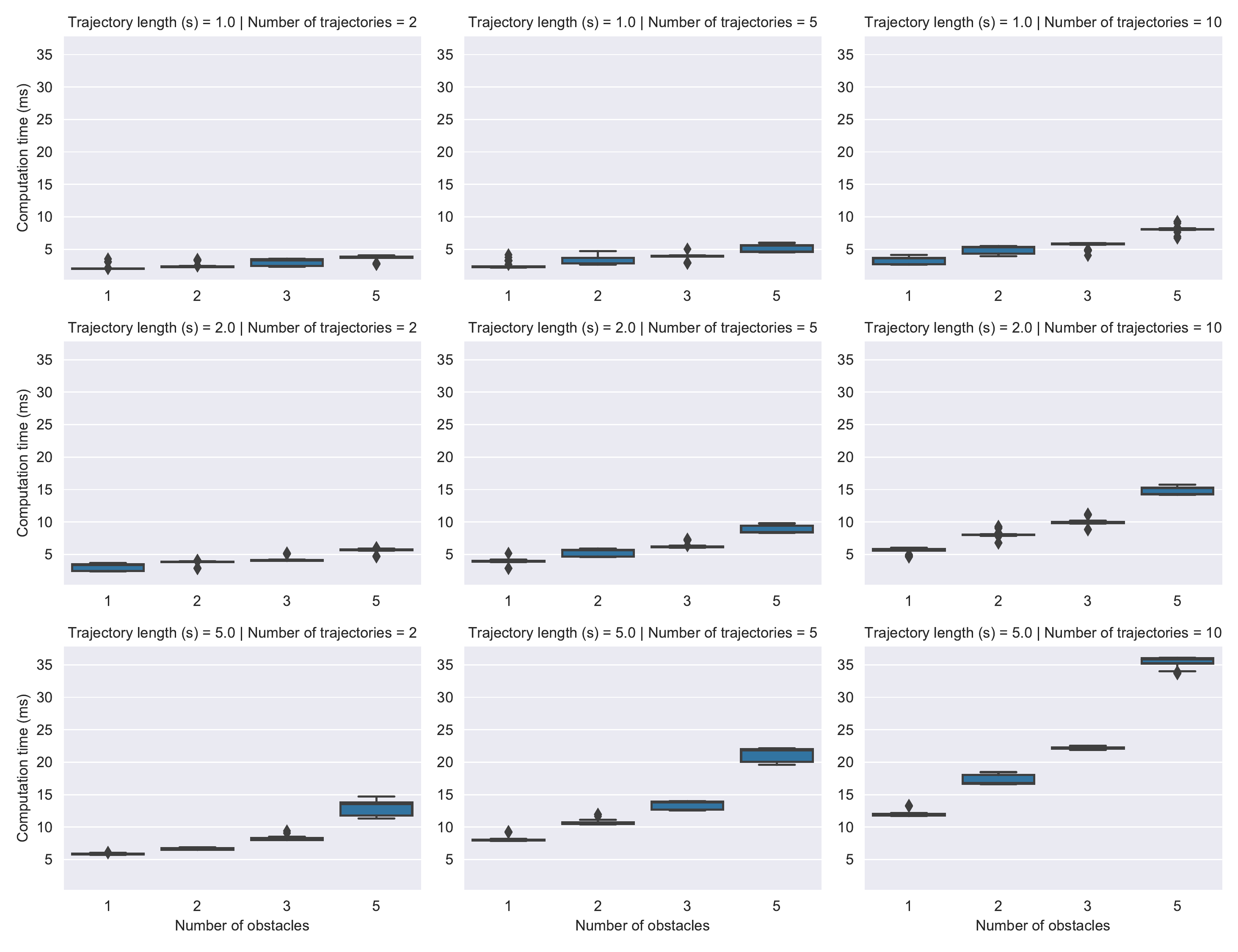}
	\caption{The computation time of the trajectory checker with respect to multiple trajectories.}
	\label{fig:ComputationTimeMultiTrajectories}
\end{figure}

Additionally, we also benchmark the performance of the trajectory checker with respect to multiple possible trajectories. As the Carla simulator provides only one trajectory for the simulated AV, instead of performing checks on different trajectories, we configure the trajectory checker to analyze the input trajectory multiple times in a sequential manner. From Figure \ref{fig:ComputationTimeMultiTrajectories}, it can be seen that, in situations with 5 obstacles, the trajectory checker can perform 10 trajectory safety checks in less than 40 milliseconds.

Notice that, since the trajectory checker is executed together with FreeRTOS, LWIP and NanoPb on a single CPU, it could be the case that the execution of those libraries increases the computation time of the trajectory checker.

\subsection{Safety assessment examples}

We now present two different safety assessment examples with results reported by the trajectory checker.

\subsubsection{Overtake and cut-in example analysis:} 
\begin{figure}[t]
	\centering
	\includegraphics[width=\linewidth]{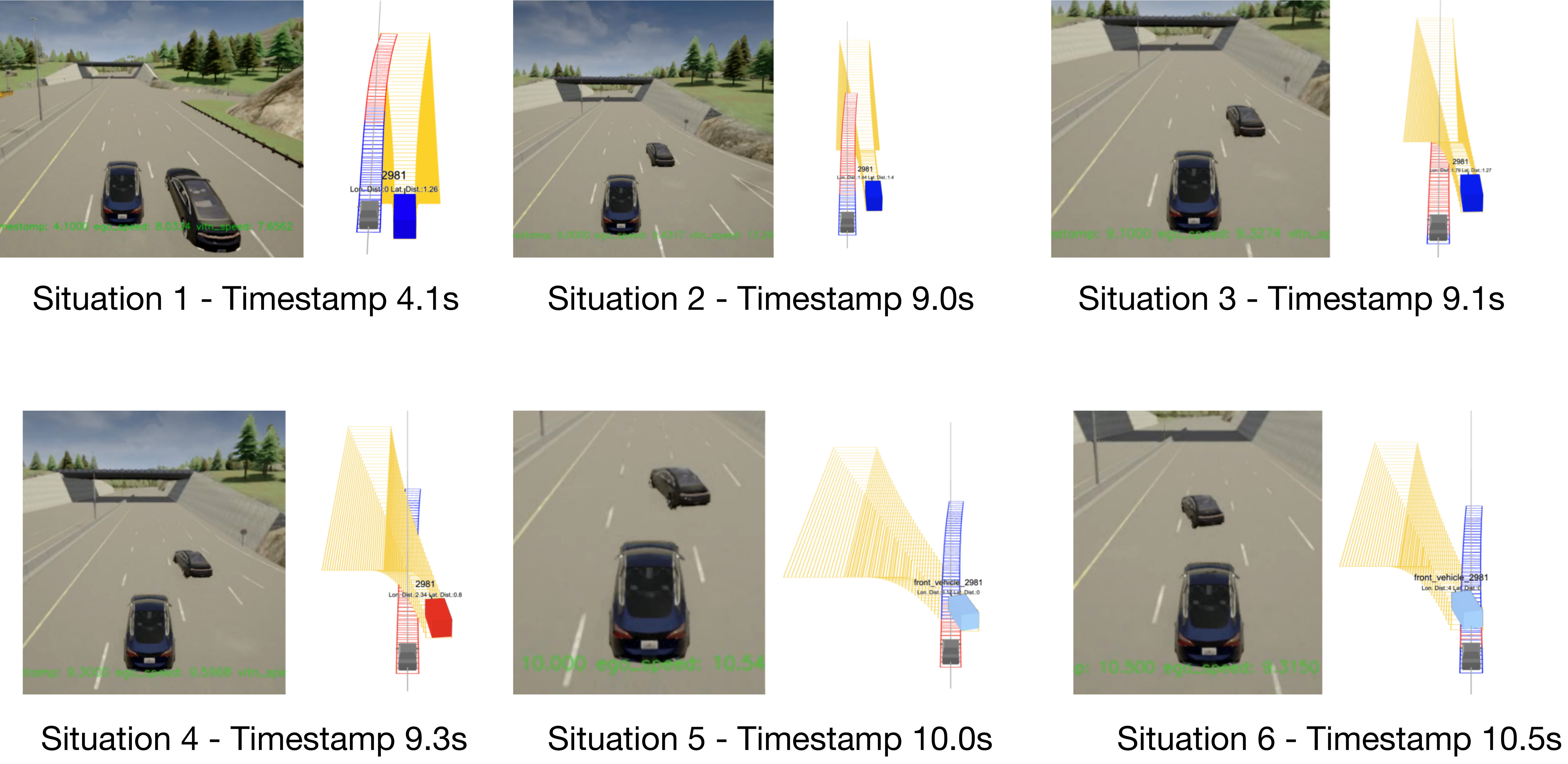}
	\caption{Safety assessment results in a cut-in scenario.}
	\label{fig:CutInResult}
\end{figure}

Our first example consists of two vehicles, the ego vehicle and a guest vehicle. In this scenario, the guest vehicle overtakes from the right and cuts-in the ego vehicle at a short distance, which poses a dangerous situation. Figure~\ref{fig:CutInResult} illustrates the safety assessment results in this scenario.

In Situation 1 at timestamp 4.1s, the guest vehicle is on the right of the ego vehicle. At this timestamp, the trajectory checker foresees that there might be a safety violation after 2.5 seconds on the trajectory of the ego vehicle if the guest vehicle performs the worst-case behavior, that is, cutting in and then braking strongly. Due to that possibility, the ego vehicle should be cautious. However, the current situation is still safe and no specific actions are required. 

Situation 2 occurs at timestamp 9.0s. Here, based on the kinematic state of the guest vehicle, the trajectory checker predicts that it will create a cut-in situation in front of the ego vehicle. The ego vehicle must be extremely cautious as there might be a safety violation in the next 1.1s if it keeps following the current trajectory and the guest vehicle brakes after cutting in.

In Situation 3 at timestamp 9.1s, the guest vehicle makes the predicted aggressive move. The trajectory checker observes that if the ego vehicle keeps its current trajectory, it will need to perform a hard brake by about $-3.7m/s^2$ in the next 0.2 seconds, which is critical. 

In Situation 4 at timestamp 9.3s, since the ego vehicle keeps following the current trajectory, without performing the observed needed brake, the trajectory checker reports that a safety constraint is violated. According to the safety policy, at this time, the ego vehicle should brake by at least $-3.64m/s^2$ but it does not.

The closer the ego vehicle gets to the guest vehicle, the higher the braking threshold that the trajectory checker reports. In this way, in Situation 5 at timestamp 10.0s, the ego vehicle is meant to brake by at least $-5.08m/s^2$ to not violate the safety rule. Since the ego vehicle does not brake more than that threshold, another violation is reported.

In this scenario, luckily, after cutting in the ego vehicle, the guest vehicle increases its speed. At timestamp 10.5s, the trajectory checker reports that given the current speed of the guest vehicle, there is no further braking requirement for the ego vehicle. However, a collision could have occurred if the guest vehicle did not speed up and the ego vehicle did not brake as required by the trajectory checker.

\subsubsection{Overtake and double cut-in example analysis:}
\begin{figure}[t]
	\centering
	\includegraphics[width=\linewidth]{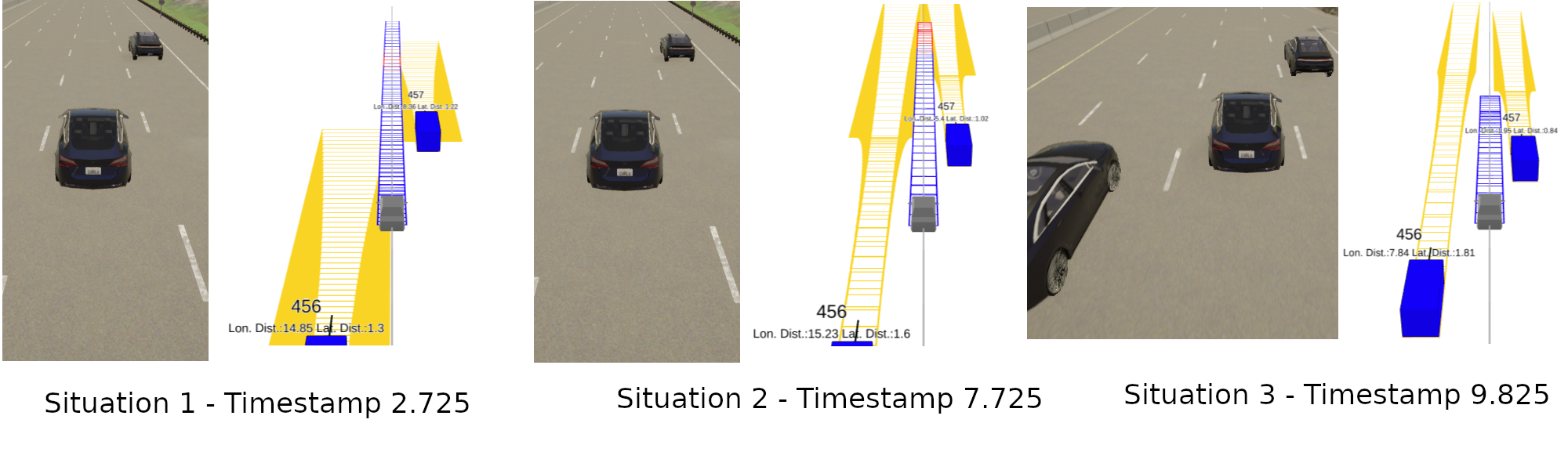}
	\caption{The safety assessment results in a two-vehicle scenario.}
	\label{fig:ConvergesResult}
\end{figure}

We additionally demonstrate our trajectory checker results in a scenario with three vehicles, accounting for the ego vehicle and two guest vehicles. In this scenario, both guest vehicles overtake and cut in the ego vehicle at short distances. This poses a dangerous situation as the ego vehicle needs to react to two dangerous maneuvers simultaneously. Figure~\ref{fig:ConvergesResult} illustrates the safety assessment results in this scenario.

Initially, the two guest vehicles are respectively on the left and on the right lane of the ego vehicle. In Situation 1 (timestamp 2.725s), the trajectory checker predicts that there might be a safety violation on the trajectory of the ego vehicle in 2.4 seconds if the guest vehicle on the right lane performs the worst-case behavior, which is cutting-in and then braking strongly. Due to that, the ego vehicle must be cautious, although it still has time to react.

In Situation 2 (timestamp 7.725s), a potential safety violation after 2.9 seconds is predicted on the trajectory of the ego vehicle if the vehicle on the right lane cuts in and then brakes strongly. In addition, a second safety violation is predicted to happen in 3.3 seconds if the vehicle on the left lane also performs the same dangerous maneuver. Although the ego vehicle needs to take into account the behavior of multiple guest vehicles at the same time, it is still safe and no actions are required.

In Situation 3 (timestamp 9.825s), the trajectory checker shows no violation since the ego vehicle decided to decrease its speed. The trajectory of the ego vehicle is now considered safe, regardless of all possible future behaviors of the guest vehicles. The lateral distances between the ego vehicle and the guest vehicles decreased rapidly over time and this induced the ego vehicle to brake so that avoiding possible collisions. In the case in which the ego vehicle would not have decelerated, the trajectory checker would have shown clearly the unsafe behavior and would have suggested the ego vehicle slow down.

\section{Conclusions}

In this paper we have demonstrated the IVEX prototype development of a trajectory checker, together with its benchmarking results on the Infineon AURIX\texttrademark TC397 platform. We have also shown and analyzed the safety assessment results reported by the prototype in two different scenarios.

The prototype is developed using many advanced techniques to guarantee the correctness of the embedded software. The validation results show the feasibility of executing the prototype on the AURIX\texttrademark TC397 microcontroller. We show that the trajectory checker can assess the safety of a 5-second trajectory in situations with 30 obstacles within less than 40 milliseconds. It can also assess the safety of 10 different trajectories with 5 obstacles within less than 40 milliseconds. The formal safety model used in the prototype will be published in a follow-up study.



\bibliographystyle{unsrt}      
\bibliography{library}   

\end{document}